# Integration of Policy and Reputation based Trust Mechanisms in e-Commerce Industry

Muhammad Yasir Siddiqui
School of Computing
Blekinge Institute of Technology
SE – 371 79 Karlskrona, Sweden

Alam Gir
School of Computing
Blekinge Institute of Technology
SE – 371 79 Karlskrona, Sweden

## ABSTRACT
The e-commerce systems are being tackled from commerce behavior and internet technologies. Therefore, trust aspect between buyer/seller transactions is a potential element which needs to be addressed in competitive e-commerce industry. The e-commerce industry is currently handling two different trust approaches. First approach consists on centralized mechanism where digital credentials/set of rules assembled, called *"Policy based trust mechanisms"*. Second approach consists on decentralized trust mechanisms where reputation/points assembled and shared, called *"Reputation based trust mechanisms"*. The difference between reputation and policy based trust mechanism will be analyzed and recommendations would be proposed to increase trust between buyer and seller in e-commerce industry. The integration of trust mechanism is proposed through mapping process, strength of one mechanism with the weakness of other. The proposed model for integrated mechanism will be presented and illustrated how the proposed model will be used in real world e-commerce industry.

## 1. INTRODUCTION
Now a day's internet has become a business hub because of its increased usage among people. E-commerce industry and online customers have been increased rapidly due to easy access. Customers to customer (C2C) e-commerce such as auction systems are more popular between individual internet users. C2C, auction systems has simple transaction process which makes this type of online shopping more popular among others. E-commerce applications are growing and getting more complex. Volume of e-commerce trading increased three times from the volume of 2007 that prevent potential users to have trust in newly arrived sellers/buyer in e-commerce industry [1].

Rapid increase of e-commerce especially for auction systems users facing more problems about trust, make it hard for new sellers and buyers to establish trustworthy relationship [1]. The current auction systems can be web applications or stand-alone software. Auction system provides ability for users to post their products for bidding. In most cases both buyer and seller don't know about each other while having a deal of transaction. From buyer aspect it's hard for buyer to trust on new seller to establishing trustworthy business partnership. Web of trust is an important area in both industry and academia. Many trust mechanisms have been developed so far, each has a different approach and characteristics about trust. Trust layer in semantic web refers to trust mechanisms which involve verification process that the source of information refers who the source claims to be and how much trustworthy it is. Verification process involves encryption and signature mechanisms that allow any consumer of this particular information to verify the source of the information. Reputation and authentication were focused according to the work previously done by different researchers on trust mechanisms [2].

## 2. BACKGROUND
Marsh was the first one who analyzed Trust as a computational concept in the distributed artificial intelligence domain [3]. Computational concepts are used currently over the web as rating systems which clearly describe positive ratings for particular web content in a particular environment. G. Zacharia proposed a model considering buyer's credit in the calculation of seller's credit, which is believed to make the evaluation more reliable. Relationship between consumer and seller while they have done transaction is the main consideration in evaluation of trust in G. Zacheria model [4].

The results of simulation indicate that it had improved effects based on G. Zacharia model in the situations like reputation collision or reputation slander [5]. Furthermore, Tale 1.0 Abdessalem described trust models and mechanisms for calculating trust between pair of users. Under his research, he explained how each participant is responsible for their ratings from other participants in distributed environment. Many social networks, e-commerce and web content systems are using rating systems such as Smart Information Systems, Smart Assistants "Based on Semantic-Web-data and is using ontology information to map customer needs to technical product attributes" [6]. Smart Information system is providing an easy way to locate data within the web trustfully. Usually reputation can be defined as the trust amount inspired by a particular person in a specific domain of interest [3]. Reputation evaluated according to its expected economic outcomes is regarded as asset creation in "Trust in a Cryptographic Economy" [7].

Another similar study was conducted by Heski Bar-Isaac on seller reputation where he introduced a framework which embeds a number of different approaches to find the seller reputation [8]. Recommended trust evaluation model is proposed by Tianhui You and Lu Li for e-commerce applications based on trust evaluation model considering the consumer's purchasing preference in e-commerce industry [9]. Tianhui model can simulate the results that indicate it had better effects, confronted with fraud behavior and trust of buyer in seller. In all the studies described above, the main focus of researchers was policy and reputation based mechanisms of trust [3][4][5][6][7][8][9].

Policy based trust mechanism has a solid authentication set of rules such as trusted certification authorities and signed certificates. Policy based trust mechanism consists on binary decisions. These decisions can be made on pre-defined policies, in response resources/services may be allowed or denied. Second trust mechanism is a reputation based which involves "soft computations" i.e. rating systems. Many rating systems are more popular over the web which are based on these reputation based trust mechanisms. Reputation based mechanisms has been more useful in semantic web or Peer-to-Peer i.e. auction systems in e-commerce industry [10]. Both policy and reputation based trust mechanisms are addressing the same problem, to establish trust between interacting parties in distributed and decentralized environment but from different perspectives and have different type of settings to act upon. Trust management will be more benefit from an intelligent integration of both policy and reputation based trust mechanisms. In some situations, trust can be better achieved from policy, while in other situations benefits may be attained by the use of reputation in such an integrated approach. An integrated mechanism will enhance the existing trust management tools and can be very effective [10].

## 2.1 Reputation based Trust Mechanisms
Personal experience and the experience of other entities in the form of ratings/feedbacks were used to make a trust decision in reputation based trust mechanism.

## 2.2 Policy based Trust Mechanisms
Policy based trust mechanism has a solid authentication set of rules such as trusted certification authorities and signed certificates. Policy based trust mechanism consists of binary decisions. These decisions can be made on the basis of given credentials by an entity, in response resources/services should be allowed or denied [10].

Usually policy and reputation based trust mechanisms are used in different organizations for trust establishment in the industry where both mechanisms have different set of rules to act upon. Policy based trust mechanism is a centralized approach where binary trust decisions has been made on some digital and logical rules. Reputation based trust mechanism as a decentralized approach where trust decision has been made on the basis of personal experience and experience of other entities i.e. rating/feedback.

In some cases trust may not be fully achieved either through policy or reputation based trust mechanism. Industry may get benefits from an intelligent integration of both policy and reputation based trust mechanisms. The purpose of this paper is to overcome weaknesses of policy and reputation based trust mechanisms by introducing an integration of both trust mechanisms.

## 3. INTEGRATION PROCESS
Reputation and Policy based trust mechanisms are being used distinctly for different environments but addressing the same situation in e-commerce industry. There is a need to integrate both trust mechanisms, which is able to fortify both trust mechanisms in order to overcome some weaknesses and get benefit of their strengths [11]. The proposed integrated trust mechanism has capabilities of both policy and reputation based trust mechanisms especially in e-commerce industry by identifying the strength and weaknesses of both trust mechanisms. Identified data can address different factors which are involved to enhance the trust using mapping process. The process of Factors identification may consist on survey, interview, literature review and expert opinions, dependent on their respected environment. Factors can be different or dependent on their own respected environments can be mapped each other directly to strengths/weaknesses of reputation and policy based trust.

## 3.1 Mapping Process
Mapping process is proposed on factors perceived from strengths and weaknesses of both trust mechanisms. Mapping processes is a design mechanism which can capture real world problems and lead towards design a solution of that particular problem [12]. For example, there are two traditional trust mechanisms on both ends one end has a weakness. To overcome this weakness, the other end's strength can be used. So engaging both ends identified weaknesses can be fulfilled with each other's strengths. Mapping of weaknesses associated with policy based trust mechanism to the strength of reputation based trust mechanism gives clear idea about the identified factors to establish trust between two entities [13].

## 3.2 Proposed Integrated Architecture
Proposed integrated trust mechanism is composed of three building blocks, the first block is based on the authentication of policy based trust mechanism and the third block is based on direct and recommended trust calculation which is directly related to reputation based trust mechanism. The real time calculation is involved in 2nd block. Second block is dependent on real time factors identified in mapping process. The designed process is based on e-commerce industry. Verified buyer and seller can participate in proposed scenario. Trust opinion is a level of trust calculated from data repository and from real time calculations which will help buyer to decide which product of a particular seller is more trusted, trust calculations are dependent on trust opinions submitted from buyer/sellers. If a seller is new in the market then policy based credential verification supports new seller to have initial trust level based on score or profile indication given by the repository. All three building blocks are explained in fig.1 [13].

One of the weakness factors was pre-registration phase asked irrelevant information which is not related to the required service. The proposed mechanism is helpful to overcome this weakness in initial verification phase. Second major weakness described in mapping section as weaknesses of policy based trust mechanisms, was the binary decisions of policy based trust mechanisms. Integration process proposed such a mechanism that could work like policy based verification and its output may similar to the reputation based trust mechanism.

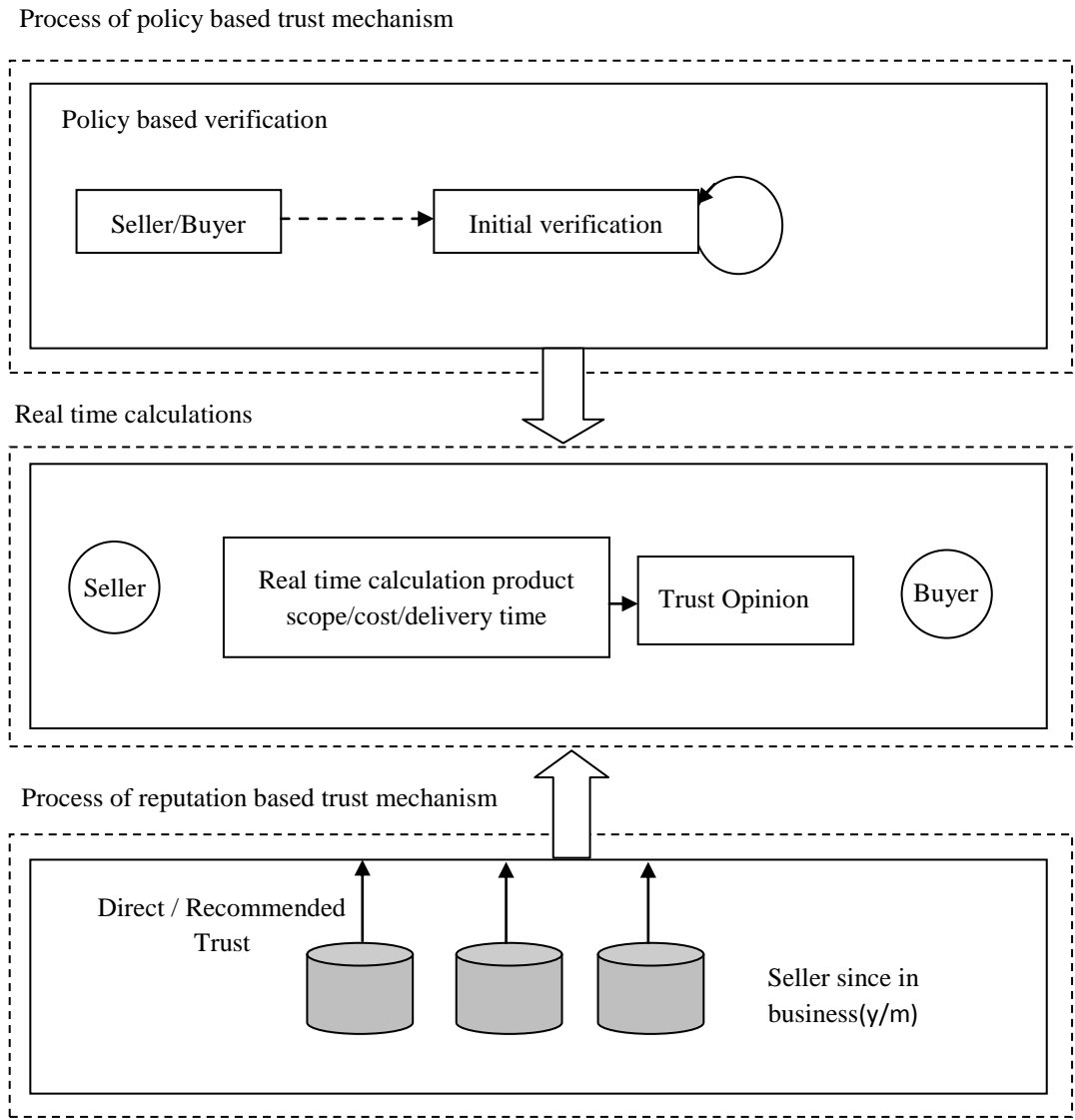

**Fig.1 Explanation of proposed integrated trust mechanism**

Proposed rating process have a cross rating where buyer/seller can rate each other after every successful transition, data repository can hold only latest rating between these two particular parties. There are two commonly used reputation calculation methods, approximate trust computation (ATC) where cache is involved to speed up the computation process. The second method is dynamic trust computation (DTC) where fresh data collection is made on runtime for further computation of trust values [14]. The proposed real time calculation is dependent on some factors identified by their own environments. In e-commerce scenario where buyers and sellers are directly involved doing transactions ranging between a few cents to thousands of dollars. There are some most important factors involved that can make an entity trustworthy or not [10]. In our proposed integrated trust mechanism the calculations are real-time based on these factors i.e. scope of product, cost of product, delivery time.

### 3.3 Policy Based Verification Process
The top layer of proposed integration process is to verify the credentials given by an entity. One of the weakness factors was pre-registration phase asked irrelevant information which is not related to the required service. Initial verification phase has the ability to overcome this weakness explained in fig.2. Second major weakness described in mapping section as weaknesses of policy based trust mechanisms, was the binary decisions of policy based trust mechanisms. Our proposed flexible proposed policy based trust mechanism has such a mechanism that could work like policy based verification and its output may similar to the reputation based trust mechanism. Suggested policy based mechanism for auction systems is described in fig.2.

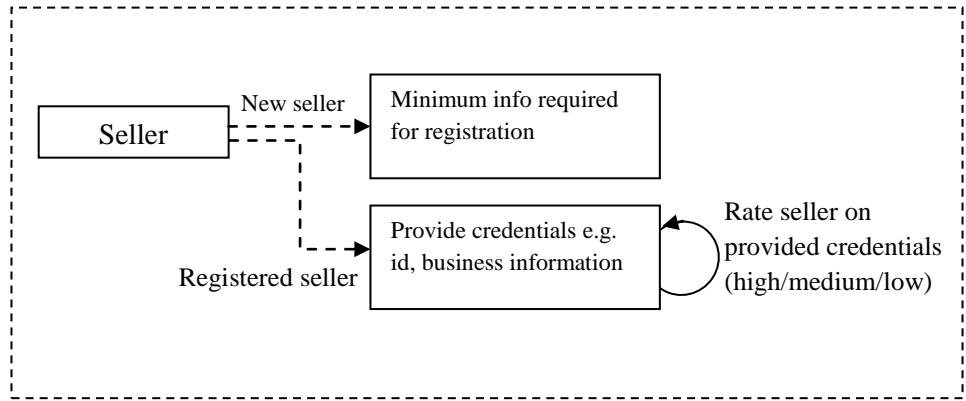

**Fig.2 First layer flexible proposed policy based trust mechanism**

## 3.4 Reputation based trust mechanism

The weaknesses of reputation based trust mechanisms fortified with the strengths of policy based trust mechanism. Identified weaknesses were fake, duplicate id and re-birth of a user addressed in suggested mapping process. Policy based credential verification is a strong shield among these kinds of weaknesses. The process of get rid from negative rating was not so simple or defined well in auction systems which is a big issue for potential seller, our proposed rating process have a capability to solve that issue as in cross rating and buyer/seller can rate each other once, data repository can hold only latest rating between two parties. Suggested detailed reputation based trust mechanism for auction systems is described in fig.3.

## 4. AN APPLICATION SCENARIO

The proposed integrated trust mechanism can be used in e-commerce industry, to enhance the trust between buyer and seller. E-commerce industry usually revolves around the purchasing and selling goods or services, the risk is involved in calculating goods and their respected shipping cost. Our proposed integrated trust mechanism has the capability to overcome these risks. An application with proposed integrated trust mechanism can be developed and tested with real world scenarios and with large amount of data. The proposed integrated trust mechanism could preferably use in cloud computing during the exchange of documents/services. Parties who are exchanging their documents or services may take trust factors into account to have trustworthy relations. Integrated approach can be more effective in medical applications, store services and with email clients.

Details of proposed reputation based trust mechanism

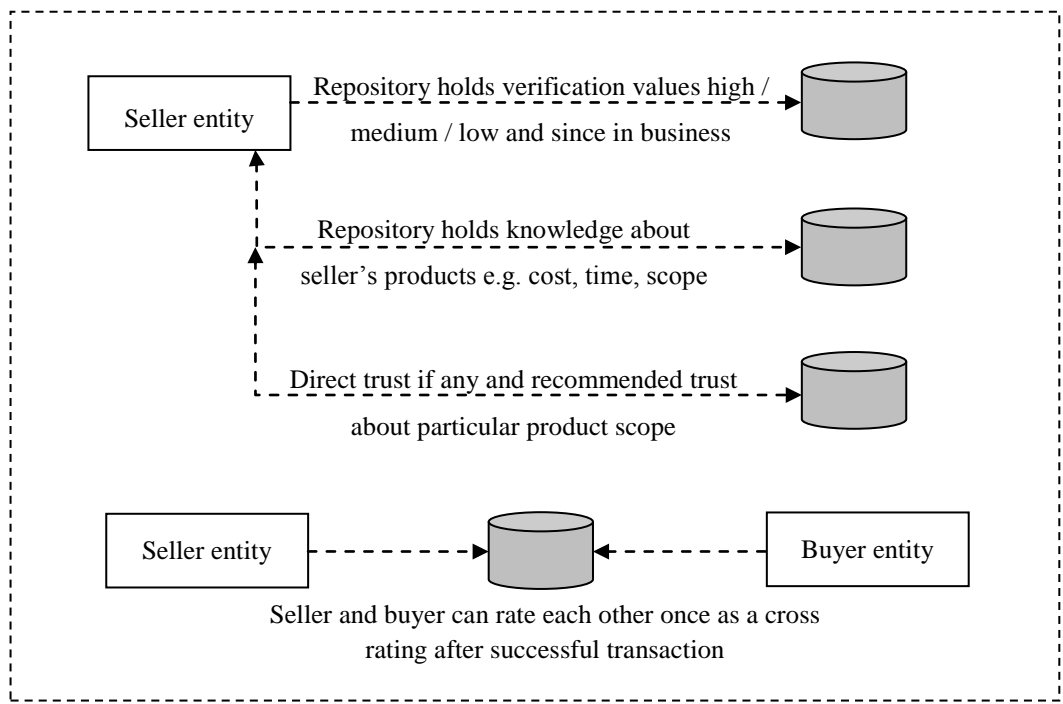

**Fig.3 Second layer detailed working of reputation based trust mechanism**

## 5. CONCLUSION

In this paper an integration of policy and reputation based trust mechanism is introduced using mapping process. Mapping process was based on identified strengths and weaknesses of both traditional trust mechanisms which were interchangeable according to their environment and needs.

There was no formal trust mechanism being used in industry that can support new seller in e-commerce industry, every e-commerce platform has their own distinct set of rules and reputation systems. The current trust mechanisms for calculating the reputation in industry is based on number of transactions, no matter what was the context of transaction and how much cost of product was involved in transaction. For these reasons an integrated trust mechanism was introduced to achieve benefit from both policy and reputation based trust mechanisms.

Although e-commerce industry is continuously improving their trust mechanisms to improve trust relationship between seller and customer but still there are chances for possible improvements where some of issues are addressed in proposed integrated trust mechanism.